# Rethinking Evidence Hierarchies in Medical Language Benchmarks: A Critical Evaluation of HealthBench

Fred Mutisya (MBChB)1, Shikoh Gitau(PhD)1, Nasubo Ongoma1, Keith Mbae1, Elizabeth Wamicha(PhD)1

Affiliation

1. [Qhala](#)   email: fname.lname@Qhala.com

*"Most clinicians in public facilities, despite being taught the ideals, have had to improvise interventions to save lives. Take, for example, a child brought to the A&E in severe dehydration. The gold standard would be to use an intraosseous line. However, most of the time, we have to improvise and use IV lines. How does the benchmark reflect this reality?"* Doctor M, Kenya

# Abstract

HealthBench, a benchmark designed to measure the capabilities of AI systems for health better (Arora et al., 2025), has advanced medical language model evaluation through physician-crafted dialogues and transparent rubrics. However, its reliance on expert opinion, rather than high-tier clinical evidence, risks codifying regional biases and individual clinician idiosyncrasies, further compounded by potential biases in automated grading systems. These limitations are particularly magnified in low- and middle-income settings, where issues like sparse neglected tropical disease coverage and region-specific guideline mismatches are prevalent.

The unique challenges of the African context, including data scarcity, inadequate infrastructure, and nascent regulatory frameworks, underscore the urgent need for more globally relevant and equitable benchmarks. To address these shortcomings, we propose anchoring reward functions in version-controlled Clinical Practice Guidelines (CPGs) that incorporate systematic reviews and GRADE evidence ratings.

Our roadmap outlines "evidence-robust" reinforcement learning via rubric-to-guideline linkage, evidence-weighted scoring, and contextual override logic, complemented by a focus on ethical considerations and the integration of delayed outcome feedback. By re-grounding rewards in rigorously vetted CPGs, while preserving HealthBench's transparency and physician engagement—we aim to foster medical language models that are not only linguistically polished but also clinically trustworthy, ethically sound, and globally relevant.

# 1. Introduction

The rapid advancements in Large Language Models (LLMs) have positioned them as transformative tools across various sectors, with healthcare emerging as a particularly high-stakes application domain. From aiding clinical decision support and patient education to facilitating telemedicine consultations, LLMs hold immense promise for enhancing efficiency and accessibility in medical practice (Cascella et al., 2023; (Elhaddad & Hamam, 2024). However, the unique challenges inherent in healthcare applications, where accuracy, safety, and trustworthiness are paramount, underscore the indispensable need for rigorous evaluation benchmarks. Traditional evaluation methods, such as multiple-choice tests, often fall short in capturing the nuanced complexities of clinical dialogue, diagnostic reasoning, and patient interaction, necessitating more sophisticated assessment tools.

## 1.1 HealthBench: A Significant Advance in Medical LLM Evaluation

The launch of HealthBench (Arora et al., 2025) represents a pivotal moment for evaluating conversational health models, marking a substantial methodological advance over previous benchmarks. Its design incorporates several commendable choices that collectively establish it as a leading yardstick for medical LLMs (Kresevic et al., 2024).

First, HealthBench distinguishes itself through its physician-driven data generation. This process involves doctors proposing high-value clinical scenarios, which are then iteratively refined by these same physicians into 5,000 polished dialogues. This approach ensures that the dataset is deeply grounded in practical clinical relevance, reflecting real-world medical interactions and challenges.

Second, the benchmark exhibits remarkable diversity in its scope. The final dataset spans 26 medical specialities, incorporates 49 languages (though with acknowledged uneven representation), and covers seven thematic categories, including critical areas such as emergency referrals and global-health constraints. This broad coverage aims to make the benchmark broadly applicable across various clinical contexts and linguistic backgrounds. However, the uneven representation highlights a persistent challenge in achieving truly equitable global coverage.

Third, HealthBench employs explicit, fine-grained rubrics for evaluation. Each dialogue is associated with approximately 11 checklist items that assign or subtract up to ±10 points across five critical behavioural axes: accuracy, completeness, context-awareness, communication quality, and instruction-following. The public availability of nearly 49,000 rubric criteria promotes transparency, allowing third parties to inspect, debate, and extend the evaluation framework.

Finally, the benchmark prioritises transparent and reproducible scoring. HealthBench is released with an open-source grader, baseline results for several frontier models, and a permissive MIT license. This open approach allows researchers and developers to re-run the benchmark or add new subsets using the provided scripts, fostering community engagement and scientific reproducibility.

Collectively, these design choices establish HealthBench as the one of the most open, physician-grounded, and comprehensive benchmarks yet released for medical LLMs (Alghamdi & Mostafa, 2024). Any critical assessment of its limitations must begin by acknowledging these significant advancements and their role in setting a new standard for evaluation.

## 1.2 The Critical Need for Evidence-Based Reward Signals

While HealthBench represents a significant leap forward in medical language model evaluation, its fundamental methodological underpinning presents a critical vulnerability: its reward signals are primarily anchored in expert opinion rather than the higher tiers of the clinical evidence pyramid. This foundational choice, while ensuring clinical realism through physician input, inadvertently introduces a scientific weakness by relying on lower-tier evidence for evaluation.

This reliance on individual expert judgment carries inherent risks, including the potential for codifying regional biases, perpetuating guideline mismatches, and embedding the idiosyncrasies of individual clinicians into the benchmark's reward structure. These issues are particularly pronounced and magnified in low- and middle-income settings (LMICs), where neglected tropical diseases (NTDs) and unique healthcare contexts remain significantly under-represented. The benchmark, despite its multi-lingual scope, implicitly risks promoting a Western-centric view of medical knowledge, making LLMs optimised on such a benchmark less effective or even potentially unsafe in diverse global contexts.

To address this fundamental challenge, a necessary evolution for medical language model evaluation involves anchoring reward functions in version-controlled Clinical Practice Guidelines (CPGs). These guidelines, which incorporate systematic reviews and GRADE evidence ratings, represent a higher standard of medical consensus and empirical validation. Such a shift promises to foster models whose outputs are not only linguistically adept but also clinically trustworthy and globally relevant, aligning AI performance with the highest available medical evidence (Huang et al., 2024).

## 2. Methodological Underpinnings and Inherent Fragilities of HealthBench

A deeper examination of HealthBench's methodological architecture reveals both its strengths in design and the inherent fragilities that arise from its reliance on expert opinion as the primary reward signal. Understanding these underpinnings is crucial for appreciating the proposed advancements.

### 2.1 Data Provenance and Rubric Authorship in HealthBench

HealthBench's dialogues are derived from three distinct streams, designed to capture a wide array of clinical interactions. The first stream consists of physician-written high-stakes scenarios, reflecting critical clinical decision points. The second involves adversarial "red-team" prompts, specifically crafted to target known blind spots or common pitfalls in medical advice, such as antibiotic misuse. The third stream transforms consumer search queries (HealthSearchQA) into structured dialogues, aiming to capture everyday patient information needs. This multi-faceted approach to data collection aims for comprehensive coverage of clinical relevance.

Regarding rubric authorship, the process is largely decentralized: a physician who either drafted or validated a specific scenario was also responsible for writing its corresponding rubric. Approximately 87% of these rubric items are attributed to a single author, with a minority undergoing double-checking via consensus. While this approach ensures clinical relevance and direct alignment with the scenario creator's intent, it also introduces a potential for individual variability in evaluation criteria.

For scalability, HealthBench employs an automated grading system powered by a GPT-4-based model. This allows for leaderboard-scale evaluation, with human review reserved for spot checks to ensure quality control. The prompts themselves are stored as role-annotated JSON lists, with each entry including tags for urgency, uncertainty, and information sufficiency. This structured data format facilitates granular control and transparency, enabling researchers to slice and analyse scores based on clinical acuity. However, these very methodological details, while offering benefits in transparency and control, also foreshadow the potential fragilities that become apparent when viewed through the lens of evidence-based medicine.

## 2.2 The Evidence Pyramid Inversion: Reliance on Expert Opinion

The core methodological concern with HealthBench lies in what can be described as an **"evidence pyramid inversion."** In classic evidence hierarchies, systematic reviews and randomized controlled trials (RCTs) occupy the apex, representing the highest quality of evidence. These are followed by consensus statements, with individual expert experience or anecdotal evidence residing at the base. HealthBench, however, inverts this order for its reward signals. A single physician's judgment, encoded within a rubric, can result in a significant penalty of ten points, effectively outweighing volumes of high-quality empirical evidence that are not explicitly incorporated into the rubric's criteria.

This reliance on individual expert opinion as the primary ground truth for evaluation is particularly problematic given established research on physician behaviour. Studies, such as that by Naumann et al. (2023), demonstrate that real-world guideline adherence varies significantly among clinicians, and even senior practitioners often deviate from established clinical guideline algorithms. This is further compounded by the presence of cognitive biases that affect human judgment, such as anchoring bias (clinging to an initial diagnosis), confirmation bias (seeking information that confirms existing beliefs), and premature closure (arriving at a conclusion too early). These thinking errors are not signs of incompetence but rather universal human tendencies in decision-making. When rubrics are predominantly single-authored or based on limited consensus, they risk codifying these individual or localized variations in practice rather than universally accepted, evidence-based guidelines. This means that a benchmark intended to promote optimal medical language model performance might inadvertently train models to adhere to a specific clinician's practice patterns, which may not be the most effective, safe, or universally applicable approach. Such an encoding of expert opinion as non-negotiable "verifiable" rewards risks penalizing nuanced, context-aware care that might, in fact, be aligned with higher-tier evidence but deviates from a specific rubric's narrow definition.

Furthermore, the automated grading mechanism, powered by a GPT-4-based model, amplifies these underlying fragilities. If the foundational rubrics are inherently flawed due to their reliance on subjective expert opinion and regional biases, then the automated grader will faithfully propagate and scale these flaws across all evaluations. This creates a scenario where the efficiency of automated grading inadvertently leads to widespread misrankings and misdirection in model development, as models are optimized against a potentially suboptimal or biased standard. The grader's potential for hallucinations or misinterpretations further compounds this issue, creating a multi-layered vulnerability that undermines the trustworthiness of the benchmark's results. For instance, the AMQA (Adversarial Medical Question-Answering) dataset, developed to benchmark bias in LLMs, revealed substantial disparities where even the least biased models answered privileged-group questions over 10 percentage points more accurately than unprivileged ones. This highlights the critical need for systematic human audit and bias detection in automated grading systems. This underscores the need for a shift from automated application of potentially flawed rules to the automated application of rigorously vetted rules, complemented by systematic human oversight.

## 2.3 Identified Limitations in Current Benchmarking Paradigms

Beyond the fundamental issue of evidence hierarchy inversion, several specific limitations within HealthBench's current design impede its ability to comprehensively and equitably evaluate medical language models for global application. These limitations, detailed below, highlight critical areas for improvement.

### 2.3.1 Sparse Coverage of Neglected Tropical Diseases (NTDs)

A significant imbalance exists in the representation of various diseases within the HealthBench corpus. While HIV which has been in existence for 40 years, and affects 1.7 Million new people, disproportionately largely in the global south it for instance, has 142 Q&A pairs, constituting approximately 2.8% of the dataset, other high-burden neglected tropical diseases (NTDs) such as malaria ( 263 million estimated cases of malaria globally in 2023) , schistosomiasis (200 Million), lymphatic filariasis ( 120 Million people, with 1.2B at risk), and trachoma ( 103 Million People)  similarly low representation. This disproportionate focus on certain conditions over others leads to a poor measurement of everyday clinical presentations prevalent in many low- and middle-income countries (LMICs).[1] This imbalance is not merely a data gap; it represents a global health equity blind spot. By implicitly privileging contexts and disease burdens more common in high-income countries, the benchmark inadvertently perpetuates a Western-centric view of medical knowledge. LLMs optimized on such a benchmark will consequently be less effective, potentially unsafe, and certainly less relevant for the vast majority of the global population residing in regions where NTDs pose significant public health challenges. Addressing this requires a fundamental shift towards equitable representation, actively seeking input from clinicians in endemic regions.

### 2.3.2 Region-Specific Guideline Mismatch

Rubrics within HealthBench often encode region-specific clinical guidelines, particularly evident in immunization schedules. For example, a direct checklist excerpt might state, "Correctly states that the influenza vaccine is given once every year, the pneumococcal vaccine is given every five years, the shingles vaccine is given as a 2-dose series for those age 50 and older, a tetanus diphtheria or Tdap booster is given every 10 years..." and penalize statements like "Incorrectly states that a flu vaccine is given twice a year to those with chronic medical conditions." However, such criteria are often aligned with US/UK immunization schedules and do not universally apply. Kenya's immunization program, for instance, omits the influenza vaccine, adult pneumococcal boosters are uncommon, and Td/Tdap is administered every pregnancy, not strictly every 10 years. Similar divergences exist in South Africa and other African countries.[1] Rewarding answers aligned to these specific rubrics risks promoting unsafe or irrelevant advice in different jurisdictions, directly impacting patient safety and clinical relevance (Chang & Ke, 2024). This underscores the need for localized and context-aware evaluation criteria.

### 2.3.3 Single-Turn Dominance

Many examples within HealthBench conclude after a single exchange between the model and the simulated patient. This design limits the benchmark's ability to assess critical aspects of conversational AI that are vital in real-world clinical interactions. Follow-up questioning, maintaining memory consistency across multiple turns, appropriate escalation or de-escalation of care, and longitudinal documentation are all critical in tele-health triage and ongoing patient management. By under-testing these multi-turn capabilities, the benchmark provides an incomplete picture of a model's clinical utility in dynamic conversational settings.

### 2.3.4 Static Snapshot Risk

Medical evidence is dynamic and constantly evolving, with new research emerging and clinical guidelines being updated frequently. However, HealthBench, as a static benchmark, is frozen in time. This poses a significant risk: scores can quickly become stale, and models may inadvertently optimise to outdated practices, such as superseded COVID-19 booster intervals. This inherent tension between the dynamic nature of medical knowledge and the static nature of traditional benchmarks means that models trained on such datasets can rapidly become clinically irrelevant or even harmful. This challenge necessitates a continuous integration/continuous deployment (CI/CD) approach to benchmark maintenance, moving towards a "living benchmark" concept that can evolve alongside medical science.

### 2.3.5 Automated GPT-4 Grader Without Systematic Human Audit

The reliance on an automated GPT-4-based grader, while enabling large-scale evaluation, introduces its own set of vulnerabilities. Grader hallucinations or misinterpretations can silently distort leaderboard rankings, leading to inaccurate assessments of model performance. Furthermore, a single AI grader can exhibit biases, potentially favoring models with similar architectures or prompting styles, thus undermining the fairness and validity of the benchmark. This points to a broader need for explainability and auditability in AI evaluation itself. If the grading mechanism operates as a black box without systematic human oversight, the trustworthiness of model improvements and leaderboard positions becomes questionable. This emphasizes that "trustworthy AI" extends beyond the model's output to the entire development and evaluation pipeline, necessitating robust human-in-the-loop validation and a "mixture of experts" approach to mitigate single-point biases. The AMQA (Adversarial Medical Question-Answering) dataset, for instance, was developed to specifically benchmark bias in LLMs, revealing substantial disparities where even the least biased models answered privileged-group questions more accurately. This highlights the critical need for systematic human audit and bias detection in automated grading systems.

The critical limitations discussed above are summarized in Table 1, along with their associated examples, implications, and proposed mitigation measures.

Table 1: Critical Limitations of HealthBench Benchmarks and Proposed Mitigations

| Issue | Example | Effects | Mitigation Measure |
|---|---|---|---|
| Sparse coverage of Neglected Tropical Diseases (NTDs) | HIV has 142 Q&A pairs (~2.8% of corpus), while malaria, schistosomiasis, lymphatic filariasis, trachoma, yaws, and | This imbalance results in poor measurement of everyday clinical presentations in many low- and middle-income | Commission clinicians from endemic regions to author 300-500 new cases per priority NTD, mirroring local guideline complexity and |
| | other high-burden NTDs have similarly low representation. | countries (LMICs).[1] | co-morbidities, and up-weight these cases during scoring until parity is achieved. |
| Region-specific guideline mismatch | Rubrics encode US/UK immunisation schedules that do not hold elsewhere. E.g., Kenya's immunization program omits influenza vaccine; Td/Tdap is administered every pregnancy, not strictly every 10 years.[1] | Rewarding answers aligned to the quoted rubric risks unsafe or irrelevant advice in different jurisdictions, directly impacting patient safety and clinical relevance.[1] | Tag rubric items with ISO country codes and load locale-specific criteria at evaluation time; include a "jurisdiction" field in prompts; cross-walk each rubric item to WHO, Kenya KEPI, South-African NDoH or other relevant guidelines and award points only if the answer matches the active guideline. |
| Single-turn dominance | Some examples end after one exchange. | Follow-up questioning, memory consistency, escalation and de-escalation—critical in tele-health triage and longitudinal documentation—are under-tested. | Ensure ≥ 50% of future dialogues contain three or more turns; add "handoff" scenarios requiring recall of earlier context. |
| Static snapshot risk | Medical evidence evolves, yet the benchmark is frozen in time (e.g., superseded COVID-19 booster intervals). | Scores can become stale; models may optimise to outdated practices. | Add metadata to tag highly variable data points (e.g., pandemic vaccinations). Release quarterly dataset versions, retire obsolete items, add new guideline changes and publish a changelog plus a "benchmark year" field for longitudinal comparison. |
| Automated GPT-4 grader without systematic human audit | Grader hallucinations or misinterpretations can silently distort leaderboard rankings. A single AI grader can be biased and favour models with similar architecture and prompting styles.[1] | Directly impacts the validity and fairness of the benchmark, potentially leading to misleading conclusions about model performance.[1] | Use a 'mixture of experts' type grader with multiple model types to prevent favouring models with a similar architecture. Audit 5–10% of scored outputs with humans each release; publish grader–human agreement statistics; maintain a public bug tracker for misgraded |

|  |  |  | examples. |
|---|---|---|---|

# 3. Anchoring Rewards in Clinical Practice Guidelines: A Proposed Framework

A promising and necessary remedy to the aforementioned fragilities is to fundamentally pivot the reward signals for medical language models from experience-based demonstrations to those rigorously grounded in vetted clinical practice guidelines (CPGs). This shift represents a crucial step towards elevating the scientific rigour and clinical trustworthiness of AI in healthcare.

## 3.1 Rationale for CPG-Grounded Reward Systems

Unlike ad-hoc physician edits or imperfect heuristics derived from electronic health records, modern CPGs are developed through a systematic and robust process. They undergo comprehensive systematic literature reviews, are assigned GRADE (Grading of Recommendations Assessment, Development and Evaluation) quality ratings based on the strength of underlying evidence, and achieve multi-stakeholder consensus. This rigorous development process positions CPGs at the higher tiers of the evidence hierarchy, making them a superior foundation for reward signals.

Anchoring rubrics to CPG statements offers several profound benefits. Firstly, it directly elevates the reward signal up the evidence hierarchy, ensuring that models are incentivised to align with the most robust and empirically supported medical knowledge. Secondly, it significantly reduces variance stemming from single-author opinions, mitigating the risk of codifying individual idiosyncrasies or regional biases. Instead, the reward system would reflect a broader, consensus-driven understanding of best practices. Thirdly, CPGs offer a public, version-controlled reference that can be systematically updated as new scientific evidence emerges and medical understanding evolves. This inherent dynamism is critical for a field as rapidly changing as medicine.

In practical terms, a guideline-anchored reinforcement learning (RL) framework would map each rubric item to a specific, citable excerpt from a CPG. This direct linkage means that when new evidence leads to an update in a guideline, the corresponding reward function can also be updated. This mechanism effectively closes the "evidence-drift gap" that is inherent in static synthetic datasets, ensuring that models are continuously optimised against the most current and validated medical knowledge. This approach moves beyond a simplistic "right/wrong" binary, fostering a living, adaptive, and ethically aware evaluation system (Freyer, Wiest, & Gilbert, 2025). It is essential to acknowledge, however, that the development and maintenance of CPGs, especially in LMICs, require sustainable funding mechanisms and significant capacity building, often necessitating additional local funding or in-kind support.

## 3.2 Roadmap for Evidence-Robust Reinforcement Learning

To achieve the vision of CPG-anchored reward systems for medical language models, a structured roadmap is essential. This roadmap outlines three key steps: developing guideline-linked datasets, implementing evidence-weighted scoring, and integrating contextual override logic.

### 3.2.1 Guideline-Linked Datasets

The objective of this foundational step is to translate narrative clinical practice guideline (CPG) statements into granular, machine-readable "reward clauses" that an AI model can be precisely scored against. This process involves several critical implementation steps:

- **Canonical mapping:** For each CPG recommendation, a persistent identifier (e.g., "WHO-Pneumonia-2023-Rec-3.2.1") is assigned. This ensures traceability and version control, allowing for precise referencing of the source of truth.
- **SMART transformation:** Narrative CPG sentences are decomposed into discrete, testable conditions. This follows the WHO Standards-based, Machine-readable, Adaptive, Requirements-based and Testable (SMART) Guidelines workflow. This transformation typically yields a FHIR Clinical Quality Language (CQL) expression alongside a natural-language checklist item, effectively bridging the gap between human-readable clinical text and machine-actionable rules. FHIR CQL is particularly suitable due to its focus on clinical quality and machine-readability, enabling precise and automated evaluation. This systematic approach allows for the integration of high-tier evidence while providing structured mechanisms for handling real-world complexities, marrying scientific rigour with practical clinical applicability.
- **Traceability ledger:** The mapping, from Guideline to Checklist to Reward clause, is stored in a version-controlled registry. This ledger is crucial for downstream audits, enabling a clear demonstration of exactly which guideline passage triggered each model reward or penalty, thereby enhancing transparency and accountability in the evaluation process.
- **User-facing esprit:** To foster trust and understanding among end-users, the numbered recommendation is surfaced alongside any model feedback. This allows clinicians to immediately trace the rationale behind a model's score or suggestion, demystifying the AI's decision-making process and promoting confidence in its outputs.

### 3.2.2 Evidence-Weighted Scoring

Not all CPG statements carry equal evidentiary weight. Recognizing this nuance, the proposed framework modulates the reward magnitude by the strength of the underlying evidence supporting each guideline recommendation. This introduces a sophisticated scoring system that reflects the certainty and robustness of medical knowledge.

The algorithmic recipe for this approach involves assigning different point values based on established evidence tiers, often referencing frameworks like GRADE. For instance:

Table 2: Evidence Tiers and Corresponding Reward Weights for CPG-Anchored Scoring

| Evidence Tier | Source (e.g., GRADE) | Weight (Δ points) | Example |
|---|---|---|---|
| High / Strong | ≥ 2 high-quality RCTs | +3 / −3 | Early antibiotics for sepsis |

|  | or meta-analysis |  |  |
| --- | --- | --- | --- |
| Moderate | 1 RCT or consistent observational data | +2 / –2 | Dexamethasone for croup |
| Low / Conditional | Single cohort, expert consensus | +1 / –1 | Zinc for the common cold |

These weights are maintained in a look-up table bundled with the checklist JSON. A critical feature of this system is its dynamic update capability: when guidelines are revised or new evidence alters a recommendation's strength, a migration script re-evaluates each clause's evidence tier and recalculates historical scores. This ensures longitudinal consistency in performance evaluation, allowing for meaningful comparisons of model capabilities over time as medical knowledge evolves. This approach acknowledges that medical knowledge is not monolithic; some recommendations are stronger than others, and this nuance is crucial for developing sophisticated medical LLMs.

### 3.2.3 Contextual Override Logic

While adherence to CPGs is paramount, real-world clinical scenarios often present complexities where strict, rigid rule-following could potentially lower patient welfare. Examples include medicine stock-outs, patient-specific contraindications, or unique resource constraints. The contextual override logic is designed to prevent undue penalties in such non-ideal scenarios, allowing for clinically appropriate deviations.

The mechanism for this involves a **dynamic rule engine**. At inference time, the language model sends its proposed plan alongside a context vector, which includes critical real-time information such as drug formulary status, patient vitals, comorbidities, and the local resource tier. This allows for an evaluation that is acutely aware of the specific clinical environment.

An **override ontology** is maintained, comprising a sanctioned set of reasons for deviation. Each reason is linked to a predefined cost-benefit profile. For example, a "β-lactam shortage" might permit a "macrolide substitute" with a minor penalty (e.g., –0.5 points) instead of a severe one (e.g., –3 points) for non-adherence. This provides a structured and transparent way to manage acceptable deviations from standard guidelines.

Crucially, every override must be accompanied by a **structured explain-and-justify mandate**. This justification, such as "Amoxicillin unavailable on Ward 7; used doxycycline per hospital policy PH-ABX-2024-14," can be presented to auditors or clinicians, ensuring accountability and transparency for deviations from standard practice.

Finally, **equity guardrails** are integrated into this system. Overrides are meticulously logged and periodically analyzed for systematic bias. This proactive monitoring aims to identify if, for instance, one demographic group consistently receives more "resource-constraint" overrides than another, thereby preventing the override mechanism from inadvertently exacerbating health inequities, particularly in low- and middle-income settings where resource disparities are common. This comprehensive framework, combining guideline-linked datasets, evidence-weighted scoring, and contextual overrides, signifies a profound shift from static, snapshot-based evaluation to a living, adaptive, and ethically-aware benchmark, capable of evolving with medical science and adapting to diverse clinical realities. It moves towards a continuous learning and evaluation paradigm, making the benchmark not just a performance metric but a critical component of ethical AI governance.

# 4. Discussion

The proposed framework for anchoring medical language model evaluation in Clinical Practice Guidelines (CPGs) represents a fundamental shift from current paradigms, offering significant implications for the development, deployment, and trustworthiness of AI in healthcare. By addressing the limitations inherent in expert-opinion-based benchmarks, this approach promises to foster models that are not only linguistically proficient but also clinically sound and globally equitable.

## 4.1 Implications for Medical LLM Development and Deployment

Anchoring reward signals in CPGs will fundamentally reshape how medical LLMs are trained and optimized. Developers will be incentivized to align their models with established, rigorously vetted medical knowledge rather than the potentially idiosyncratic opinions of individual experts. This will lead to more consistent, reliable, and predictable model outputs, which is paramount in a high-stakes domain like medicine. The ability to provide transparent, evidence-backed rationales for a model's suggestions, facilitated by the "user-facing esprit" of guideline-linked datasets, directly addresses the "black box" nature often associated with AI. This enhanced transparency and explainability will foster greater trust among clinicians, patients, and regulatory bodies, which is crucial for the responsible adoption and widespread deployment of LLMs in diverse healthcare settings, from clinical decision support systems to patient-facing applications. This paradigm shift signals to developers that "good" medical AI is not merely about linguistic fluency or factual recall, but about clinical trustworthiness, global applicability, and ethical consciousness, aligning with ethical frameworks that emphasize fairness, privacy, and transparency (Singhal, 2024), such as those adopted by leading healthcare companies like Johnson & Johnson.

## 4.2 Addressing Global Health Equity and Contextual Nuances

The proposed framework directly confronts the critical issues of sparse Neglected Tropical Disease (NTD) coverage and region-specific guideline mismatches that plague current benchmarks. By advocating for the commissioning of clinicians from endemic regions to author new cases and implementing locale-specific evaluation criteria, the framework ensures a more equitable representation of global disease burdens and diverse clinical practices. This moves beyond a Western-centric view, making LLMs more relevant and effective for the vast majority of the world's population.

Benchmarks like Alama Health-QA, which is anchored on Kenyan Clinical Practice Guidelines and demonstrates high coverage of NTDs, are essential for safe and equitable model evaluation and deployment across African health systems. Similarly, AfriMed-QA, by incorporating specific diseases and

Qhala 11

local challenges prevalent in Africa and guiding models towards feasible treatment recommendations, directly addresses the need for culturally sensitive and contextually relevant AI. The "Contextual Override Logic" is vital for enabling adaptive care in real-world, often resource-constrained environments or in situations involving patient-specific contraindications. This mechanism allows models to navigate the complexities of medical practice where strict guideline adherence might not be optimal or even feasible, moving beyond a rigid, one-size-fits-all approach to healthcare. The inclusion of "Equity guardrails" within this logic is particularly significant. By systematically logging and analyzing overrides for potential biases, the framework proactively monitors and mitigates the risk of inadvertently perpetuating or exacerbating health disparities, especially in low- and middle-income countries. This comprehensive approach ensures that the benchmark itself becomes a tool for promoting global health equity and responsible AI deployment.

## 4.4 Challenges and Future Directions

While the proposed framework offers a robust path forward, its implementation is not without challenges. A significant effort will be required for the data acquisition and curation necessary to translate vast amounts of narrative CPGs into granular, machine-readable "reward clauses." This demands robust infrastructure, specialised expertise in medical informatics, and extensive collaborative efforts across clinical, technical, and public health domains. The severe lack of systematic and well-structured health data, with only 1% originating from African countries, remains a major hurdle.

Interoperability and standardization also pose a challenge. Achieving seamless integration of CPGs into benchmarks will necessitate standardizing their formats, potentially through widespread adoption of standards like FHIR Clinical Quality Language (CQL), across different healthcare organizations and national guidelines. Furthermore, Africa's computational infrastructure is often insufficient, and internet penetration is low in many regions, limiting the viability of AI-based health systems. Prioritizing investments in digital and computational infrastructure is essential to create a robust foundation for AI applications.

Looking ahead, extending the framework to evaluate more complex clinical reasoning beyond mere checklist-based accuracy will be crucial. This includes assessing a model's capability in diagnostic reasoning, generating differential diagnoses, understanding disease progression, and formulating long-term care plans, aspects that may not be fully captured by current CPGs alone.

**Ethical considerations** remain paramount and require ongoing vigilance. Issues such as data privacy (concerns about patient information handling exist in countries like Kenya and South Africa), the potential for algorithmic bias (even with evidence-based anchoring), and accountability for AI-driven clinical decisions must be continuously addressed throughout the development and deployment lifecycle. The framework's emphasis on transparency and auditability is a step in this direction. Still, continuous research and policy development are needed, especially given the absence of robust regulatory frameworks in many African countries. Specific challenges include data re-identification, data ownership, and the lack of laws on the use of AI in developing countries.

Finally, the future role of **human-AI collaboration** will be critical. Clinicians will remain indispensable in auditing, refining, and overseeing both the CPG-anchored benchmarks and the LLMs themselves. This emphasizes a collaborative rather than purely autonomous paradigm for medical AI. As a potential future enhancement, integrating **delayed outcome feedback**—such as patient health outcomes, readmission rates, or disease progression—into the reward function could complement CPG-based rewards, moving towards a more direct assessment of real-world impact and clinical utility. However, collecting such data in LMICs presents challenges due to complex hospital systems, documentation issues, limited awareness, financial constraints, and privacy concerns. This holistic approach will ensure that benchmarks evolve into strategic tools for driving responsible AI innovation, prioritizing safety, fairness, and clinical utility over raw performance metrics. Efforts like the Africa CDC's 'AI for Health in Africa' strategy aim to cover product life cycles, ethics, and data governance for developing sophisticated, ethical, and culturally sensitive AI systems for African health applications, alongside plans for scaled-up development of specialized AI-trained workforces through improved STEM education (Amugongo et al., 2025; Schmitt, 2022). This is particularly crucial given that the Government AI Readiness Index 2021 ranked Africa as one of the regions with the lowest level of readiness for AI adoption (Maslej et al., 2025).

# 5. Conclusion

HealthBench has undeniably set a new standard for evaluating conversational medical language models, leveraging physician-driven data generation and transparent scoring to create a robust yardstick. However, its foundational reliance on expert opinion as the primary reward signal introduces critical fragilities, particularly concerning global relevance, the dynamic nature of medical knowledge, and the potential for embedding biases. This approach risks codifying regional idiosyncrasies and penalizing clinically appropriate, evidence-based care that deviates from a single expert's view.

To address these shortcomings, this paper proposes a comprehensive framework for anchoring reward functions in systematically developed Clinical Practice Guidelines (CPGs). This involves creating guideline-linked datasets through canonical mapping and SMART transformation, implementing evidence-weighted scoring to reflect the strength of underlying evidence, and incorporating contextual override logic to account for real-world clinical complexities and resource constraints. Such a framework moves beyond a static, subjective evaluation to a living, adaptive, and ethically-aware benchmark system.

The integration of an African context is paramount, acknowledging the unique disease burdens, healthcare infrastructure challenges, and the need for culturally and contextually relevant AI solutions. Benchmarks like Alama Health-QA and AfriMed-QA exemplify the critical role of regionally curated, guideline-aligned datasets in fostering LLMs that are effective and safe for diverse African populations. By re-grounding rewards in rigorously vetted CPGs—while diligently preserving HealthBench's commendable transparency and physician engagement—we can foster the development of medical language models whose outputs are not only linguistically polished and coherent but also clinically trustworthy, globally relevant, and ethically sound. This evolution is essential for realising the full potential of AI in healthcare, ultimately contributing to safer, more equitable, and more effective healthcare delivery worldwide.

large-scale study of relevance assessments with large language models: An initial look. arXiv preprint arXiv:2411.08275.

Zhang, K., Yang, X., Wang, Y., Yu, Y., Huang, N., Li, G., ... & Yang, S. (2025). Artificial intelligence in drug development. Nature medicine, 31(1), 45-59.

Xiao, Y., Huang, J., He, R., Xiao, J., Mousavi, M. R., Liu, Y., Li, K., Chen, Z., & Zhang, J. M. (2025). AMQA: An Adversarial Dataset for Benchmarking Bias of LLMs in Medicine and Healthcare. *arXiv preprint arXiv:2505.19562*.

Yagos, A., et al. (2024). Transforming African Healthcare with AI: Paving the Way for Improved Health Outcomes. *Journal of Collaborative Healthcare and Translational Medicine*.(https://www.jscimedcentral.com/jounal-article-info/Journal-of-Collaborative-Healthcare-and-Translational-Medicine/Transforming-African-Healthcare-with-AI:-Paving-the-Way-for-Improved-Health-Outcomes-11760)

Qhala

17